\documentclass{svproc}
\usepackage{graphicx}

\usepackage{times}  
\usepackage{helvet}  
\usepackage{courier}  
\usepackage{graphicx} 
\usepackage{caption} 
\usepackage{algorithm}
\usepackage{algorithmic}

\usepackage{booktabs}
\usepackage{bm}
\usepackage{makecell}
\usepackage{multirow}
\usepackage{amsfonts}
\usepackage{amsmath}
\usepackage{tabularx}
\usepackage{newfloat}
\usepackage{listings}

\usepackage{url}

\begin{document}
\mainmatter              
\title{HCL: Improving Graph Representation with Hierarchical Contrastive Learning}
\titlerunning{HCL: Improving Graph Representation with Hierarchical Contrastive Learning}  
%


\author{Jun Wang\inst{1}\thanks{Equal Contribution.}
\and
Weixun Li\inst{1\star} \and
Changyu Hou\inst{1} \and
Xin Tang\inst{2} \and
Yixuan Qiao\inst{1} \and
Rui Fang\inst{2} \and
Pengyong Li\inst{3}\thanks{Corresponding Author.} \and
Peng Gao\inst{1} \and
Guotong Xie\inst{1,4,5\star\star}
}

%
\authorrunning{ }
%
\institute{Ping An Healthcare Technology, Beijing, China \and
Ping An Property \& Casualty Insurance Company, Shenzhen, China \and
School of Computer Science and Technology, Xidian University, Xian, China \and
Ping An Health Cloud Company Limited, Shenzhen, China \and
Ping An International Smart City Technology Company Limited, Shenzhen, China
}


\maketitle              

\begin{abstract}
Contrastive learning has emerged as a powerful tool for graph representation learning. However, most contrastive learning methods learn features of graphs with fixed coarse-grained scale, which might underestimate either local or global information. To capture more hierarchical and richer representation, we propose a novel Hierarchical Contrastive Learning (HCL) framework that explicitly learns graph representation in a hierarchical manner. Specifically, HCL includes two key components: a novel adaptive Learning to Pool (L2Pool) method to construct more reasonable multi-scale graph topology for more comprehensive contrastive objective, a novel multi-channel pseudo-siamese network to further enable more expressive learning of mutual information within each scale. Comprehensive experimental results show HCL achieves competitive performance on 12 datasets involving node classification, node clustering and graph classification. In addition, the visualization of learned representation reveals that HCL successfully captures meaningful characteristics of graphs.
\end{abstract}

\section{Introduction}
\noindent Graph representation learning has recently attracted increasing research attention, because of broader demands on exploiting ubiquitous non-Euclidean graph data across various domains, including social networks, physics, and bioinformatics~\cite{hamilton_2017_nips}. Along with the rapid development of graph neural networks (GNNs) ~\cite{kipf2016semi,hamilton_2017_nips}, GNNs have been reported as a powerful tool for learning expressive representation for various graph-related tasks. However, supervised training of GNNs usually requires faithful and labour-intensive annotations and relies on domain expert knowledge, which hinders GNNs from being adopted in 
practical applications.


Self-supervised learning has emerged as a powerful tool to alleviate the need for large labelled data. Among them, contrastive learning has recently achieved promising results~\cite{hassani2020contrastive}.
Contrastive learning techniques are used to train an encoder that builds discriminative representations by comparing positive and negative samples to maximize the mutual information (MI)~\cite{liu2020self}. 


Although the graph contrastive learning GCL methods have achieved significant success, they suffer all or partially from the following limitations. First, most contrastive learning methods like DGI~\cite{velickovic_2019_dgi}, GCA~\cite{zhu2021gca}, and GRACE~\cite{grace}, learn features of graphs with fixed fine-grained scale, which might underestimate either local or global information. However, each graph has multi-scale intrinsic structures, including the grouping of nodes into motifs, the further grouping of motifs into sub-graphs as well as the spatial layout of sub-graphs in the topology space. Such multi-scale intrinsic structures are more flexible and informative, and can provide important clues for graph representation learning. In most cases, a single level contrastive objective could merely capture limited characteristics of graphs~\cite{velickovic_2019_dgi,zhu2021gca,grace}. Second, considering that existing GCL methods heavily rely on negative samples to avoid representation collapse, To alleviate this limitation, Grill et al.~\cite{grill2020bootstrap} propose the Bootstrap Your Own Latent (BYOL) framework to perform unsupervised representation learning on images by leveraging the bootstrapping mechanism with Siamese networks\cite{simsiam}. However, Siamese networks have not been well extended to graph domain yet. We argue that bootstrapping graphs with a multi-channel scheme would enable graph encoders to capture more powerful representation. 



To address the aforementioned limitations, we propose a novel Hierarchical Contrastive Learning (HCL) framework, HCL constructs a cross-scale contrastive learning mechanism to learn hierarchical graph representation in an unsupervised manner. More specifically, the two key components of HCL including: (i) a Learning to Pool (L2Pool) method with topology-enhanced self-attention to recursively construct a series of coarser graphs during multi-scale contrastive learning and (ii) a contrastive objective term that preserves the mutual information with expressive multi-channel networks. The simple yet powerful framework can be optimized in an end-to-end manner to capture more comprehensive graph features for downstream tasks. 
To summarize, this work makes the following major contributions:

\begin{figure*}[!h]
\centering
\includegraphics[width=0.99\textwidth]{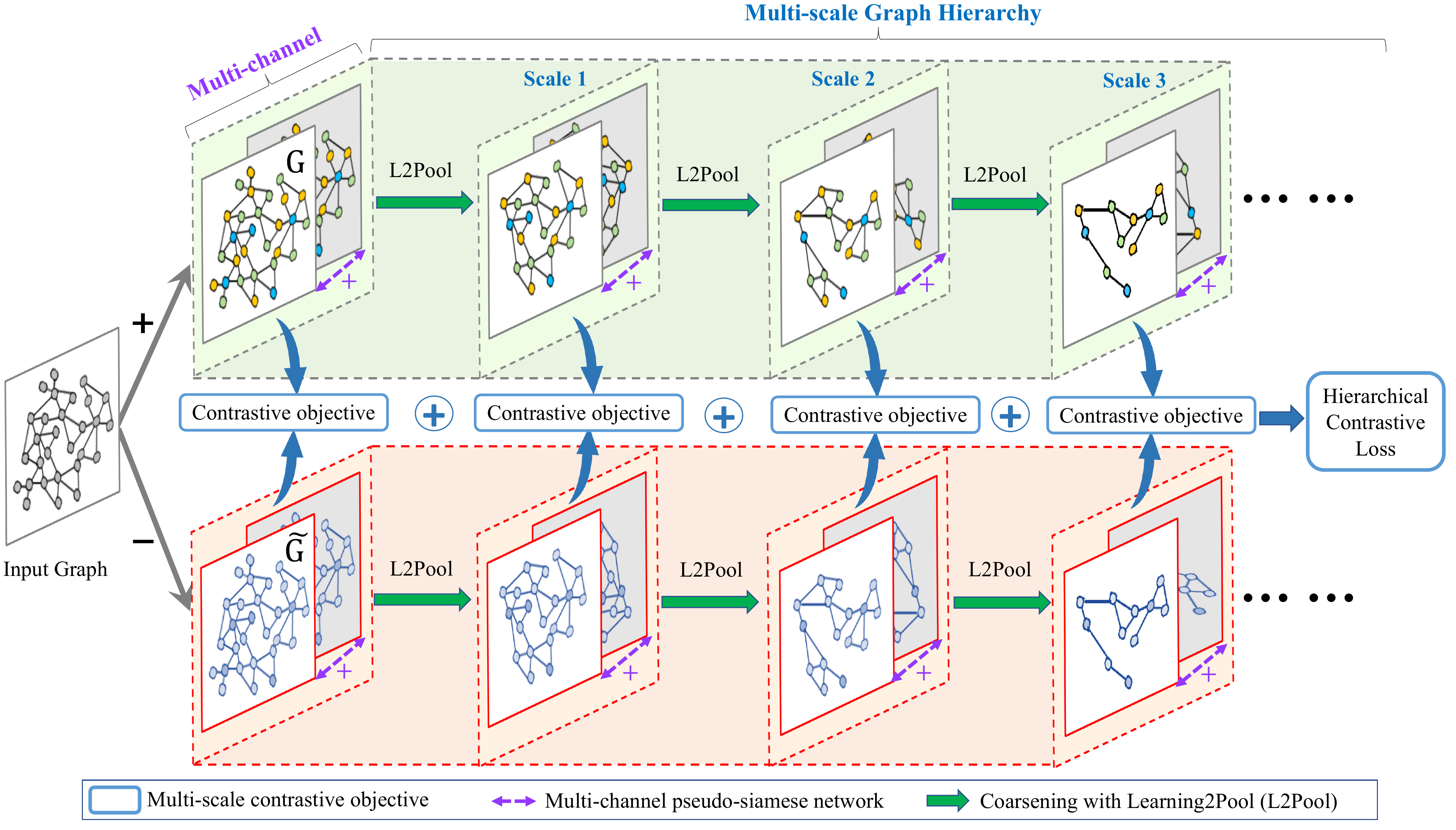}
\caption{Framework of the proposed Hierarchical Contrastive Learning (HCL) for graph representation.}
\label{fig_framework}
\end{figure*}

\begin{itemize}

\item We propose a novel Hierarchical Contrastive Learning (HCL) framework to learn graph representation by taking advantage of hierarchical MI maximization across scales and bootstrapping multi-channel contrastiveness across networks.

\item We proposed a novel L2Pool method to form fine to coarse-grained graph and contrastive objective across scales, which explicitly preserves information concealed in the hierarchical topology of the graph.

\item Extensive experiments indicate that HCL achieves superior or comparable results on various real-world 12 benchmarks involving both node-level and graph-level tasks. Moreover, visualization of nodes representation further reveals that HCL can capture more intrinsic patterns underlying the graph structures. 

\end{itemize}

\section{Related Works}

\subsection{Unsupervised Graph Learning}
Traditional graph unsupervised learning methods are mainly based on graph kernel~\cite{narayanan2017graph2vec}. 
Compared to graph kernel, contrastive learning methods can learn explicit embedding, and achieve better performance, which are the current state-of-the-art for unsupervised node and graph classification tasks~\cite{gmi,hassani2020contrastive}. Generally, current contrastive graph learning employs a node-node contrast~\cite{zhu2021gca,gmi} or node-graph contrast~\cite{velickovic_2019_dgi,hassani2020contrastive} to maximize the mutual information at single level.
For example, DGI~\cite{velickovic_2019_dgi} employs the idea of Deep InfoMax~\cite{hjelm_2019_iclr} and consider both patch and global information during the discrimination. MVGRL~\cite{hassani2020contrastive} introduces augmented views to graph contrastive learning and optimizes the DGI-like objectives. 
Besides, GRACE~\cite{grace}, InfoGraph~\cite{sun2020infograph} and SUBG-CON~\cite{jiao2020sub}, further extend the idea of graph MI maximization and conduct the discrimination across the node, sub-graph and graph. PHD~\cite{ijcai2021-phd} using graph-graph contrast reports impressive performances on graph classification, but not for the node-level tasks. Nevertheless, most of them contrast graphs with fixed scales, which might underestimate either local or global information. To address these issues, our HCL explicitly formulates multi-scale contrastive learning on graphs and enables capturing more comprehensive features for downstream tasks.

\subsection{Multi-scale Graph Pooling } 
Early graph pooling methods use naive summarization to pool all the nodes~\cite{gilmer_2017_icml}, and usually fail to capture graph topology. Recently, multi-scale pooling methods have been proposed to address the limitations. Among them, graph-coarsening pooling methods like DiffPool~\cite{ying2018diffpool} and StructPool~\cite{yuan2020structpool} consider pooling as a node clustering problem, but the high computational complexity of these methods prevents them from being applied to large graphs. On the other hand, the node-selection pooling methods like gPool~\cite{gao2019graph} and SAGPool~\cite{lee2019self} preserve representative nodes based on their importance, but tend to lose the original graph structures. Compared to previous works, the proposed HCL has two main differences: \textbf{1)} Apart from the common late fusion of features, HCL uses L2Pool and Pseudo-siamese network to intermediately aggregate richer contrastive objectives across scales, where the embeddings at various scales in each network layer are fused to enable richer contrasting in a hierarchical manner. \textbf{2)} The proposed L2Pool module is trained given an explicit optimization for node selection with topology-enhanced Transformer-style attention, hence effectively coarsen the original graph structure. 


\section{Methodology}

\subsection{Overview}

The goal of HCL is to provide a framework to construct a multi-scale contrastive scheme that incorporate inherent hierarchical structures of the data to generate expressive graph representation. In this section, we introduce HCL and its main components in Figure \ref{fig_framework}. 
First, given an input graph $\mathbf{G}(\mathbf{X},{\bf A})$ with node features, $\mathbf{X}\in\mathbb{R}^{N \times d}$, ${\bf A}$ is the adjacency matrix. We first generate positive (green) and negative (red) samples by attribute shuffling~\cite{velickovic_2019_dgi}. Specifically, We perform the row-wise shuffling on the feature matrix $X$, so the negative graph consists of the same nodes as the original graph, but they are located in different places in the graph, and therefore receive different contextual information.
Second, for the positive branch above and the negative branch below, we both learn graph representations at multiple scales. 
We first employ a graph propagation layer on the input graph to initially embed the original scale of graph as $G_0(\mathbf{X}_0, \mathbf{A}_{0})$ with $\mathbf{X}_0 = \mathbf{X}$, $\mathbf{A}_{0} = {\bf A}$, where the graph propagation layer is implemented as a multi-channel pseudo-siamese network, with each channel using a graph convolution layer of the same structure but different weights~\cite{kipf2016semi}. We then recursively apply L2Pool for $S$ times to obtain a series of coarser scales of graph $G_1(\mathbf{X}_1, \mathbf{A}_1), \dots, G_S(\mathbf{X}_S, \mathbf{A}_S)$ 
where $|\mathbf{X}_{s}|>|\mathbf{X}_{s'}|$ for $\forall~ 1 \leq s < s' \leq S$.
Thirdly, we learn the parameters through optimizing the fused multi-scale and multi-channel contrastive loss function. 
During the inference, we 
take the graph adjacency as inputs for downstream tasks.

To train our model end-to-end and learn multi-scale representation for downstream tasks, we jointly leverage cross-scale contrastive loss. Specifically, the overall objective function is defined as:

\begin{equation}
    \mathcal{L_{}} = \mathcal{L}_0 +
    \sum_{k'=1}^k{ ((\prod_{k'=1}^{k}\alpha_{p_{k'}}) * \mathcal{L}_{p_{k'}})} \ ,
    \label{eql1}
\end{equation}

where $\mathcal{L}_0$ is the contrastive loss at the first scale with all nodes, $k$ is the total number of pool layers besides $\mathcal{L}_0$. The ${\alpha_{p_{k'}}}$ is the pooling ratio of ${k'-th}$ pooling scale, e.g., 0.9, etc. Then, ${\mathcal{L}_{p_{k'}}}$ is contrastive loss at ${k'-th}$ pooling scale.

\begin{figure}[!ht]
\centering
\includegraphics[width=0.7\columnwidth]{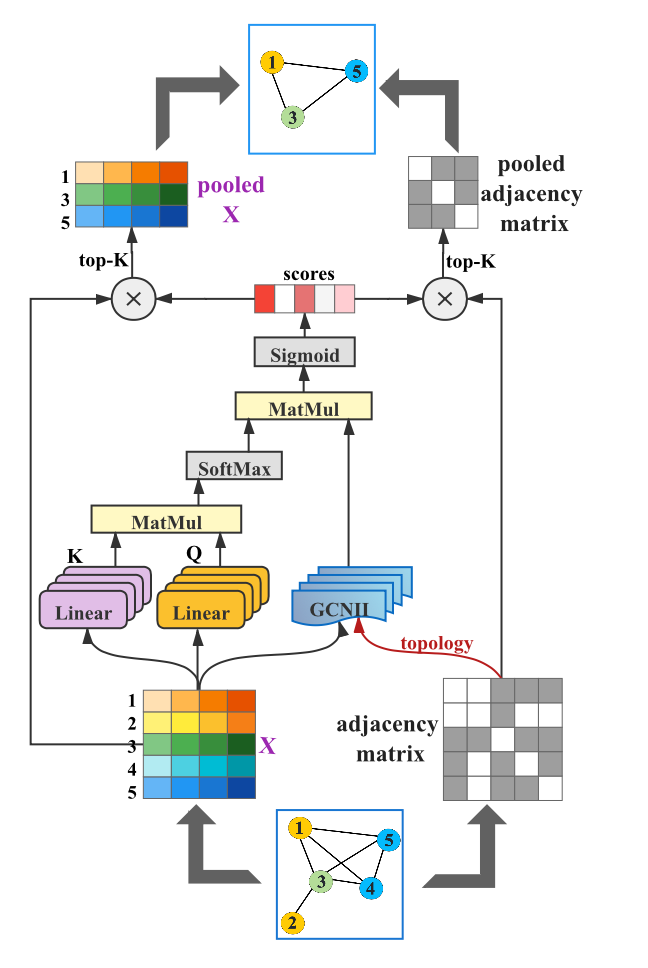}
\vspace{-0.5cm}
\caption{An illustration of the proposed L2Pool using Transformer-style self-attention and topology information to select representative nodes and to coarsen into a graph hierarchy for cross-scale contrastive learning. 
} \label{fig:gpool}
\end{figure}

\subsection{Multi-Scale Contrasting with L2Pool}

In this section, in order to create graph contrasting at multiple scales, we propose a novel Learning to Pool method, namely L2Pool, to enable coarsening graph data and contrasting information interchange across scales explicitly. L2Pool adaptively creates graph representations at multiple scales, by selecting a subset of nodes to form a new but smaller graph with topology-enhanced attention. 

As shown in Figure \ref{fig:gpool}, we implement a Transformer-style multi-head (MH) attention mechanism. 
While MH self-attention is superior to trivial pooling methods such as sum or mean, as it considers global dependencies among nodes. Moreover, note that for each node, the self-attention only calculates the semantic similarity between current node and other nodes, without considering the structural information of a graph reflected on the nodes and the relation between node pairs. To tackle this limitation, we define a novel multi-head attention enhanced with topological structure from GCNII~\cite{chen2020simplegcnii}. Specifically, GCNII is a GCN model with two effective techniques: Initial residual and Identity mapping, GCNII relieves the problem of over-smoothing thus enables deeper networks. 
The input of the attention function (Att) consists of query $Q \in \mathbb{R}^{{n_q}\times{d_k}}$, key $K \in \mathbb{R}^{{n}\times{d_k}} $ and value $V \in \mathbb{R}^{{n}\times{d_v}} $, where $n_q$ is the number of query vectors, $n$ is the number of input nodes, $d_k$ is the dimension of the key vector, and $d_v$ is the dimension of the value vector. Then we compute the dot product of the query with all keys, to put more weights on the relevant values, namely nodes, as follows: ${\rm Att}(Q,K,V)=\sigma(QK^T)V$, where $\sigma$ is an activation function. 
The output of the multi-head attention function can be formulated as:


\begin{small}
\begin{equation}
\begin{aligned}
\label{eq:kqv}
 {\rm MH}(Q,K,V) &=[O_1,...,O_h]W^o,\\
 O_i &={\rm Att}(QW^Q_i, KW^K_i, VW^V_i), \\
  &={\rm Att}(QW^Q_i, KW^K_i, {\rm GCNII}^V_i(H, A)),\\
\end{aligned}
\end{equation}
\end{small}

where 
the learning parameter matrices corresponding to $Q$, $K$ and $V$ are $W^Q_i\in \mathbb{R}^{d_k\times d_k}$, $W^K_i \in \mathbb{R}^{d_k\times d_k}$, and $W^V_i\in \mathbb{R}^{d_v \times d_v}$ respectively. Also, the output projection matrix is  $W^O \in \mathbb{R}^{d_v \times d_{model}}$, where $d_{model}$ is the output dimension for the multi-head attention function.

More specifically, we construct $V$ using GCNII, to explicitly leverage the global structure and capture the interaction between nodes according to their structural dependencies.
The multi-head self-attention enhanced by graph topology is defined as:


\begin{small}
\begin{equation}
\begin{aligned}
    {\rm GCNII}(H, A) =
    \sigma(((1-\alpha ){
    A}H+{\alpha }H^0)((1-{\beta})I_n+{\beta }W)), \\
  {\rm Att}(Q, K, {\rm GCNII}(H, A)) = {\rm softmax}(\frac{{QK^T}}{{\sqrt{d_k}}}){\rm GCNII}(H, A), \\
\end{aligned}\label{eq:transformer}
\end{equation}
\end{small}

 where $\alpha$ and $\beta$ are hyperparameters and $I_n$ is the identity matrix. Formally, given node embeddings $H \in \mathbb{R}^{n \times d}$  with their adjacency information $A$, we construct the value $V$ using a 4-layer GCNII, to explicitly leverage the graph topology information (the equation for a single layer GCNII is given in above equation 3).

Specifically, we named the learnable score function as L2Pool at layer $l$, and select the high scored nodes $i^{(l+1)} \in \mathbb{R}^{n_{l+1}}$, to drop the unnecessary nodes, denoted as follows:
\begin{equation}
    y^{(l)} = {\rm L2Pool}({\rm Att}, H^{(l)}, A^{(l)}); \  \ i^{(l+1)} = \text{top}_{k}(y^{(l)}),
\end{equation}

where $\text{top}_{k}$ function samples the top k nodes by dropping nodes with low scores $y^{(l)} \in \mathbb{R}^{n_l}$. In this way, HCL could preserve as much information as possible from the graph hierarchy and contrast in a multi-scale manner.

\subsection{In-scale Bootstrapping Pseudo-Siamese Network}

In HCL, we introduce a Pseudo-Siamese architecture to form the basic bootstrapping contrastiveness with multi-channel. Generally, the siamese network contains two identical subnetworks has been proved to be a common structure in unsupervised visual representation learning~\cite{simsiam}, but not been well extended to graph domain yet. Hence, we make a Pseudo-Siamese network with non-weight-sharing branches for multi-channel contrastive learning, which provides more flexibility and capacity than a restricted siamese network.

Inspired by above contrastive scheme, we train the GNN-encoder $f_{GNN}$ to maximize the mutual information (MI) between node (fine-grain) representations, i.e., $\mathbf{H} = f_{GNN}(\mathbf{X},\mathbf{A})$, and a global representation (summary of all representations).  This encourages the encoder to prefer the information that is shared across all nodes. 
Since maximizing the precise value of mutual information is intractable, thus, a Jensen-Shannon MI estimator is often used~\cite{hjelm_2019_iclr,oord2018representation}, which maximizes MI's lower bound. The Jensen-Shannon-based estimator acts like a standard binary cross-entropy (BCE) loss, whose objective maximizes the expected $\log$-ratio of the samples from the joint distribution (positive examples) and the product of marginal distributions (negative examples). The positive examples are pairings of $\mathbf{s}$ with $\mathbf{h}_i$ of the real input graph  $ \mathbf{G} = (\mathbf{X},\mathbf{A})$, but the negatives  are pairings of $\mathbf{s}$ with $\tilde{\mathbf{h}}_i$, which are obtained from a fake/generated input graph $ \tilde{\mathbf{G}} =(\tilde{\mathbf{X}},\tilde{\mathbf{A}})$ with $\tilde{\mathbf{H}} = f_{GNN}(\tilde{\mathbf{X}},\tilde{\mathbf{A}})$. Then, a discriminator $\mathcal{D}_1: \mathbb{R}^{F'} \times \mathbb{R}^{F'} \to \mathbb{R}$ is used to assign higher scores to the positive examples than the negatives, as in~\cite{hjelm_2019_iclr,oord2018representation}.
The Jensen-Shannon-based BCE objective with weighted sum of multi-channels across networks in ${k-th}$ pooling scale is expressed as:


\begin{small}
\begin{equation}
    \begin{split}
         \mathcal{L}_{p_k} & = \sum_{u=1}^N \mathbb{E}_{(\bm{X}, \bm{A})} \Big[ \log \mathcal{D}_{p_k}(\bm{h}^{(1)}_u + \bm{h}^{(2)}_u * \mathbf{\delta}_{p_k}, \bm{s}) \Big] \\
        & + \sum_{v=1}^N \mathbb{E}_{(\tilde{\bm{X}}, \tilde{\bm{A}})} \Big[ \log \big(1 - \mathcal{D}_{p_k}( \tilde{\bm{h}}^{(1)}_v + \tilde{\bm{h}}^{(2)}_v * \mathbf{\delta}_{p_k}, \bm{s}) \big) \Big]  \ ,\\
    \end{split}
    \label{eql2}
\end{equation}
\end{small}

with $\bm{A} \in \mathbb{R}^{N \times N}$ and $\bm{X} \in \mathbb{R}^{N \times F}$, for simplicity. $\bm{h^{(1)}_u}$ and $\bm{h^{(2)}_u}$ represent the embedding of the first channel and the second channel of the pseudo siamese network, respectively. 
We use the average function over all node features to obtain the entire graph representation, $\bm s={\rm READOUT}(X_{p_k})$ is the summary vector represents the embedding of $k-th$ pooled graph.
$\mathbf{\delta}_{p_k}$ is the weighted sum parameter between multi-channels in the $k-th$ pooling scale. This approach effectively maximizes mutual information between summary vector $\bm s$ and ${\bm{h}}^{(1)}_u + {\bm{h}}^{(2)}_u * \mathbf{\delta}_{p_k}$ in every pooling layer.

\begin{table}[!h]
\centering

\resizebox{0.75\columnwidth}{!}{%
\begin{tabular}{ccccccc}
\toprule
 & \textbf{Dataset} & \textbf{Graphs} & \textbf{Nodes} & \textbf{Edges} & \textbf{Features} & \textbf{Classes}\\
\midrule
\multirow{7}{*}{\rotatebox{90}{Node-level}} & Cora & 1 & 2,708 & 5,429 & 1,433 & 7\\
& Citeseer & 1 & 3,327 & 4,732 & 3,703 & 6\\
& Pubmed & 1 & 19,717 & 44,338 & 500 & 3\\
& Amazon-C & 1 & 13,752 & 245,861 & 767 & 10\\
& Amazon-P & 1 & 7,650 & 119,081 & 745 & 8\\
& Coauthor-CS & 1 & 18,333 & 81,894 & 6,805 & 15\\
& Coauthor-Phy & 1 & 34,493 & 247,962 & 8,415 & 5\\
\midrule 

\multirow{5}{*}{\rotatebox{90}{Graph-level}} & IMDB-B & 1,000 & 19.77 & 193.06 & - & 2\\
& IMDB-M & 1,500 & 13.00 & 65.93 & - & 3\\
& PTC-MR & 344 & 14.29 & 14.69 & - & 2\\
& MUTAG & 188 & 17.93 & 19.79 & - & 2\\
& Reddit-B & 2,000 & 508.52 & 497.75 & - & 2\\
\bottomrule
\end{tabular}
}

\vspace{0.5cm}

\caption{The statistics of the datasets.}
\label{table:dataset}
\end{table}

\section{Experiments}

In this section, we describe the experiments conducted to demonstrate the efficacy of proposed HCL for graph representation tasks. 
The experiments aim to answer the following five research questions: 
\begin{itemize}
\item \textbf{RQ1.} 
    How does HCL perform in node-level graph representation tasks? 
\item \textbf{RQ2.} 
 How does HCL perform in graph-level  representation tasks? 
\item \textbf{RQ3.} 
    How does the hierarchical mutual information maximization mechanism improve the performance of HCL? 
\item \textbf{RQ4.} 
    How do the difference parameter settings influence the performance of HCL? 

\item \textbf{RQ5.} 
    Does HCL capture meaningful patterns and provide insightful representation? 
\end{itemize}

\subsection{Datasets and Experimental Setup}

\textbf{Datasets.} We evaluate the quality of learned node and graph embeddings on downstream tasks. According to the tasks, seven of them are utilized for node-level tasks, include node classification and clustering, while five of them are for graph-level classification task.
Statistics of datasets used are shown in Table~\ref{table:dataset}. For \textbf{node classification}, we adopt 3 citation networks including Cora, Citeseer, Pubmed~\cite{sen2008collective}, and 4 co-purchase and co-author networks including Amazon-Computers, Amazon-Photo, Coauthor-CS and Coauthor-Phy~\cite{shchur2018pitfalls}. For \textbf{node clustering}, we adopt three benchmark datasets: Cora, Citeseer and Pubmed~\cite{sen2008collective}. For \textbf{graph classification}, we use another five common datasets: MUTAG, PTC-MR~\cite{chen2007chemdb}, IMDB-B, IMDB-M and REDDIT-B~\cite{yanardag_2015_kdd}.

\vspace{0.5cm}

\noindent \textbf{Experimental setup. } 
 We initialize the parameters using Xavier initialization \cite{glorot2010understanding} and train the model using Adam optimizer with an initial learning rate of 0.001 and an NVIDIA V100 GPU with 16G memory. For multi-channel configuration,
the weight sum parameter $\delta$ is learned between -1 and 1. To have fair comparisons, we set the size of the hidden dimension of both node and graph representations to 512. Specifically, HCL has set up a total of 3 recursive pooling scales of 0.9-0.8-0.7, which preserves 90\%(0.9), 72\%(0.9*0.8) to 50.4\%(0.9*0.8*0.7) nodes from the original graph, respectively. In the construction of multi-scale graphs, L2Pool is implemented with 4 attention heads and a 4-layer GCNII.  \textbf{1) For node classification tasks}, we follow DGI~\cite{velickovic_2019_dgi} to use same GCN encoder for all methods, and report the mean classification accuracy with standard deviation on the test nodes after 50 runs of training followed by a linear model. On citation networks, we use the same training/validation/testing splits as~\cite{yang_2016_icml} for training the classifier according to the node representations. Specifically, we use 20 labelled nodes per class as the training set, 20 nodes per class as the validation set, and the rest as the testing set. On co-purchase and co-author networks, we use 30 labelled nodes per class as the training set, 30 nodes per class as the validation set, and the rest as the testing set. For a fair comparison, the performances of all the methods are obtained on the same splits. The mean classification accuracy with standard deviation on the test nodes after 50 runs of training is reported. 
\textbf{2) For node clustering tasks}, we employ k-means on the obtained node representations, the clustering results averaged over 50 runs in terms of NMI and ARI are reported.
\textbf{3) For graph classification tasks}, we follow InfoGraph~\cite{sun2020infograph} to fairly evaluate the performances of HCL. The graph embedding was obtained by averaging all embedding of nodes in the graph. The mean 10-fold cross validation accuracy with standard deviation after 5 runs followed by a linear SVM is reported. We follow InfoGraph to choose the number of GCN layers, number of epochs, batch size, and the C parameter of the SVM from $[2, 4, 8, 12]$, $[10, 20, 40, 100]$, $[32, 64, 128, 256]$, and  $[{10^{-3}}, {10^{-2}}, ..., {10^2}, {10^3}]$, respectively. The parameters of classifiers are independently tuned using cross validation on training folds of data, and the best average classification accuracy is reported for each method.

\begin{table*}[!h]
\centering
\resizebox{1.0\columnwidth}{!}{%
\begin{tabular}{llccccccc}
\hline
\textbf{Method} & \textbf{Input} & \textbf{Cora}       & \textbf{Citeseer}   & \textbf{Pubmed}     & \textbf{Amazon-C}   & \textbf{Amazon-P}   & \textbf{Coauthor CS} & \textbf{Coauthor Phy} \\ \hline
MLP             & X,Y            & 58.2 ± 2.1          & 59.1 ± 2.3          & 70.0 ± 2.1          & 44.9 ± 5.8          & 69.6 ± 3.8          & 88.3 ± 0.7           & 88.9 ± 1.1            \\
LogReg          & X,A,Y          & 57.1 ± 2.3          & 61.0 ± 2.2          & 64.1 ± 3.1          & 64.1 ± 5.7          & 73.0 ± 6.5          & 86.4 ± 0.9           & 86.7 ± 1.5            \\
LP              & A,Y            & 68.0                  & 45.3                & 63.0                  & 70.8 ± 0.0          & 67.8 ± 0.0          & 74.3 ± 0.0           & 90.2 ± 0.5            \\
Chebyshev       & X,A,Y          & 81.2                & 69.8                & 74.4                & 62.6 ± 0.0          & 74.3 ± 0.0          & 91.5 ± 0.0           & 92.1 ± 0.3            \\
GCN             & X,A,Y          & 81.5                & 70.3                & 79.0                  & 76.3 ± 0.5          & 87.3 ± 1.0          & \underline{91.8 ± 0.1}           & 92.6 ± 0.7            \\
GAT             & X,A,Y          & \underline{83.0 ± 0.7}          & \underline{72.5 ± 0.7}          & \underline{79.0 ± 0.3}          & 79.3 ± 1.1          & 86.2 ± 1.5          & 90.5 ± 0.7           & 91.3 ± 0.6            \\
SGC             & X,A,Y          & 81.0 ± 0.0          & 71.9 ± 0.1          & 78.9 ± 0.0          & 74.4 ± 0.1          & 86.4 ± 0.0          & 91.0 ± 0.0           & 90.2 ± 0.4            \\
MoNet           & X,A,Y          & 81.3 ± 1.3          & 71.2 ± 2.0          & 78.6 ± 2.3          & \underline{83.5 ± 2.2}          & \underline{91.2 ± 1.3}          & 90.8 ± 0.6           & \underline{92.5 ± 0.9}            \\ \hline
DGI             & X,A            & 81.7 ± 0.6          & 71.5 ± 0.7          & 76.9 ± 0.5          & 75.9 ± 0.6          & 83.1 ± 0.5          & 90.0 ± 0.3           & 91.3 ± 0.4            \\
GMI             & X,A            & 80.9 ± 0.7          & 71.1 ± 0.2          & 78.0 ± 1.0          & 76.8 ± 0.1          & 85.1 ± 0.1          & 91.0 ± 0.0           & OOM                   \\
GRACE           & X,A            & 80.0 ± 0.4          & \underline{71.7 ± 0.6}          & \underline{79.5 ± 1.1}          & 71.8 ± 0.4          & 81.8 ± 1.0          & 90.1 ± 0.8           & 92.3 ± 0.6            \\
SUBG-CON        & X,A            & \underline{82.5 ± 0.3}          & 70.9 ± 0.3          & 73.13 ± 0.5         & OOM                 & OOM                 & OOM                  & OOM                   \\
GCA             & X,A            & 80.5 ± 0.5          & 71.3 ± 0.4          & 78.6 ± 0.6          & \underline{80.8 ± 0.4}          & \underline{87.1 ± 1.0}          & \textbf{\underline{91.3 ± 0.4}}           & \underline{93.1 ± 0.3}            \\

MVGRL & X,A   & 82.0 ± 0.7 & 70.7 ± 0.7 & 74.0 ± 0.3 & 76.2 ± 0.6 & 84.1 ± 0.3 & 83.6 ± 0.3 & 87.1 ± 0.2 \\

\textbf{HCL(Ours)} & X,A   & \textbf{82.5 ± 0.6} & \textbf{72.0 ± 0.5} & \textbf{79.2 ± 0.6} & \textbf{84.0 ± 0.7} & \textbf{87.5 ± 0.4} & {91.1 ± 0.4} & \textbf{93.3 ± 0.5} \\
 \hline
GCA*             & X,D            & 81.8 ± 0.8          & 72.0 ± 0.5          & {81.2 ± 0.7}          & {81.5 ± 0.9}          & 87.0 ± 1.2          & {91.6 ± 0.7}           & {93.0 ± 0.5}            \\
MVGRL* & X,D   & 82.8 ± 1.0 & 72.7 ± 0.5 & 79.6 ± 0.8 & 82.9 ± 0.9 & 86.9 ± 0.5 & 91.0 ± 0.6 & 93.2 ± 1.0 \\
\textbf{HCL(Ours)*} & X,D   & 
{\textbf{83.7 ± 0.7}} & 
{\textbf{73.3 ± 0.4}} & 
{\textbf{81.8 ± 0.7}} & 
{\textbf{83.4 ± 0.5}} & 
{\textbf{87.3 ± 0.4}} & 
{\textbf{91.7 ± 0.3}} & 
{\textbf{93.5 ± 0.4}} \\
 \hline
\end{tabular}
}

\vspace{0.5cm}

\caption{Node classification accuracies (\%) for supervised and unsupervised methods on different datasets. The best performance is highlighted in bold. The previous best performance is underlined. The Input column highlights the data available to each model during the model training process (X:features, A:adjacency matrix, D:diffusion matrix, Y:labels). * denotes model using Diffusion instead of Adjacency matrix as input. OOM indicates Out-Of-Memory on a 16GB GPU. Some results without standard deviations are directly taken from~\cite{hassani2020contrastive}.}
\label{table:node_class}

\end{table*}

\subsection{Evaluation on node-level tasks (RQ1)}
\noindent\textbf{Node Classification. }
To evaluate node classification under the linear evaluation protocol, we compare results of our HCL with recent unsupervised models in Table \ref{table:node_class}, including DGI~\cite{velickovic_2019_dgi}, GMI~\cite{gmi}, MVGRL~\cite{hassani2020contrastive}, GRACE~\cite{grace}, GCA~\cite{zhu2021gca}and SubG-CON \cite{jiao2020sub}.
Moreover, we also compare our results with supervised models including MLP, Logistic Regression(LogReg), label propagation (LP) \cite{zhu_2003_icml}, Chebyshev \cite{defferrard_2016_nips}, GCN, GAT~\cite{velickovic_2018_gat}, SGC~\cite{sgc} and mixture model networks (MoNet) \cite{Monti_2017_cvpr}. The results show that our HCL achieves superior performances with respect to previous unsupervised models. For example, on Amazon-C dataset, we achieve 84.0\% accuracy, which is a 3.1\% relative improvement over previous state-of-the-art. 
Furthermore, inspired by MVGRL~\cite{hassani2020contrastive}, employing Diffusion matrices other than Adjacency matrices has been shown to improve GNNs performance~\cite{klicpera_2019_nips}. We also conducted experiments of HCL with Diffusion matrices $D$ as input. Noting that, HCL with $X$ and diffusion matrix $D$ as input further yields even better performances than that of ($X,A$). HCL also outperforms both GCA and MVGRL using diffusion matrix in the same settings, which further denotes the superiority of HCL.  
 

\vspace{0.5cm}

\noindent\textbf{Node Clustering. }
To evaluate performance on node clustering task, we compare our HCL with models reported including: variational GAE
(VGAE) \cite{kipf_2016_arxiv}, marginalized GAE (MGAE) \cite {wang_2017_cikm}, adversarially regularized GAE (ARGA) and VGAE (ARVGA) \cite{pan_2018_ijcai}, GALA \cite{park_2019_iccv} and MVGRL~\cite{hassani2020contrastive}. The results in Table \ref{table:node_cluster} suggest that our model achieves superior or comparable performance on NMI and ARI scores across most of the benchmarks.
Besides, the improvements are more significant in terms of ARI compared to those of NMI. The results encourage that unsupervised clustering task prefers the representation containing the important and semantic feature due to the lack of supervised information. Meanwhile, HCL boosts the supervised classification with a larger margin, by adequately exploiting the labels and graph inherent characteristics. 
Thus, HCL tends to capture faithful and comprehensive information of the graph by enhancing the scheme of message passing.

\begin{table}[ht]
\centering
\resizebox{0.7\columnwidth}{!}{%
\begin{tabular}{lccccccllllll}
\cline{1-7}
\multirow{2}{*}{\textbf{Method}} & \multicolumn{2}{c}{\textbf{Cora}} & \multicolumn{2}{c}{\textbf{Citeseer}} & \multicolumn{2}{c}{\textbf{Pubmed}} \\ \cline{2-7}
                                 & \textbf{NMI}    & \textbf{ARI}    & \textbf{NMI}      & \textbf{ARI}      & \textbf{NMI} & \textbf{ARI}         \\ \cline{1-7}
K-means                          & 0.321           & 0.230           & 0.305             & 0.279             & 0.001        & 0.002                \\
Spectral                         & 0.127           & 0.031           & 0.056             & 0.010             & 0.042        & 0.002                \\
BigClam                          & 0.007           & 0.001           & 0.036             & 0.007             & 0.006        & 0.003                \\
GraphEncoder                     & 0.109           & 0.006           & 0.033             & 0.010             & 0.209        & 0.184                \\
DeepWalk                         & 0.327           & 0.243           & 0.088             & 0.092             & 0.279        & 0.299                \\
\cline{1-7}
GAE                              & 0.429           & 0.347           & 0.176             & 0.124             & 0.277        & 0.279                \\
VGAE                             & 0.436           & 0.346           & 0.156             & 0.093             & 0.229        & 0.213                \\
MGAE                             & 0.511           & 0.445           & 0.412             & 0.414             & 0.282        & 0.248                \\
ARGA                             & 0.449           & 0.352           & 0.350             & 0.341             & 0.276        & 0.291                \\
ARVGA                            & 0.450           & 0.374           & 0.261             & 0.245             & 0.117        & 0.078                \\
GALA                             & 0.577           & 0.531           & 0.441             & 0.446             & 0.327        & 0.321                \\
MVGRL                            & 0.572           & 0.495           & 0.469    & \textbf{0.449}    & 0.322        & 0.296                \\
\cline{1-7}
\textbf{HCL(Ours)}                    & \textbf{0.586}  & \textbf{0.536}  & \textbf{0.472}    & 0.447    & \textbf{0.332}    &
\textbf{0.329}    &
\multicolumn{1}{l}{} \\ \cline{1-7}
\end{tabular}
}

\vspace{0.5cm}

\caption{Performance on node clustering task reported in normalized MI (NMI) and adjusted rand index (ARI) measures. The best performance is highlighted in bold.}
\label{table:node_cluster}

\end{table}

\begin{table}[!bt]
\centering
\renewcommand\arraystretch{1.2}
\resizebox{0.9\columnwidth}{!}{%
\begin{tabular}{cccccccccccc}
\cline{1-7}
                              & \textbf{Method}      & \textbf{MUTAG}       & \textbf{PTC-MR}      & \textbf{IMDB-B}    & \textbf{IMDB-M}  & \textbf{REDDIT-B}  \\ \cline{1-7}
\multirow{5}{*}{\rotatebox{90}{KERNEL}}  
& SP  & 85.2 ± 2.4 & 58.2 ± 2.4  & 55.6 ± 0.2 & 38.0 ± 0.3 & 64.1 ± 0.1\\
& GK & 81.7 ± 2.1 & 57.3 ± 1.4  & 65.9 ± 1.0 & 43.9 ± 0.4 & 77.3 ± 0.2\\
& WL & 80.7 ± 3.0 & 58.0 ± 0.5  & 72.3 ± 3.4 & 47.0 ± 0.5 & 68.8 ± 0.4\\
& DGK & 87.4 ± 2.7 & 60.1 ± 2.6  & 67.0 ± 0.6 & 44.6 ± 0.5 & 78.0 ± 0.4\\
& MLG & 87.9 ± 1.6 & 63.3 ± 1.5 & 66.6 ± 0.3 & 41.2 ± 0.0 & $-$\\
\cline{1-7}
\multirow{5}{*}{\rotatebox{90}{SUPERVISED}}   & GraphSAGE            & 85.1 ± 7.6           & 63.9 ± 7.7           & 72.3 ± 5.3           & 50.9 ± 2.2           & OOM                    \\
                              & GCN                  & 85.6 ± 5.8           & 64.2 ± 4.3           & 74.0 ± 3.4           & 51.9 ± 3.8           & 50.0 ± 0.0           \\
                              & GIN-0                & 89.4 ± 5.6           & 64.6 ± 7.0           & 75.1 ± 5.1           & 52.3 ± 2.8           & 92.4 ± 2.5           \\
                              & GIN-$\epsilon$               & 89.0 ± 6.0           & 63.7 ± 8.2           & 74.3 ± 5.1           & 52.1 ± 3.6           & 92.2 ± 2.3           \\
                              & GAT                  & 89.4 ± 6.1           & 66.7 ± 5.1           & 70.5 ± 2.3           & 47.8 ± 3.1           & 85.2 ± 3.3           \\ \cline{1-7}
\multirow{8}{*}{\rotatebox{90}{UNSUPERVISED}} & random walk          & 83.7 ± 1.5           & 57.9 ± 1.3           & 50.7 ± 0.3           & 34.7 ± 0.2           & OOM                    \\
                              & node2vec             & 72.6 ± 10.2          & 58.6 ± 8.0           & OOM                    & OOM                    & OOM                    \\
                              & sub2vec              & 61.1 ± 15.8          & 60.0 ± 6.4           & 55.3 ± 1.5           & 36.7 ± 0.8           & 71.5 ± 0.4           \\
                              & graph2vec            & 83.2 ± 9.6           & 60.2 ± 6.9           & 71.1 ± 0.5           & 50.4 ± 0.9           & 75.8 ± 1.0           \\
                              & Infograph            & 89.0 ± 1.1           & 61.7 ± 1.4           & 73.0 ± 0.9           & 49.7 ± 0.5           & 82.5 ± 1.4           \\
                              

                            & GCC                & {86.4 ± 0.5}           & 58.4 ± 1.2           & {--}          & --           & 88.4 ± 0.3           \\
                              & GraphCL              & 86.8 ± 1.3           & OOM                    & 71.1 ± 0.4           & OOM                    & 89.5 ± 0.8           \\
                              & MVGRL                & \textbf{89.7 ± 1.1}           & 62.5 ± 1.7           & {74.2 ± 0.7 }          & 51.2 ± 0.5           & 84.5 ± 0.6           \\
\cline{2-7}
                              & \textbf{HCL(Ours)}                 &      {89.2 ± 1.2}  & \textbf{63.1 ± 1.4}  & \textbf{74.3 ± 0.6}  & \textbf{52.0 ± 0.6}  & \textbf{91.9 ± 0.7}  \\ \cline{1-7}
\end{tabular}
}
\vspace{0.5cm}
\caption{Mean 10-fold cross validation accuracies (\%) on graph classification task. The best performance is highlighted in bold.}
\label{tab: graph_class}

\end{table}


\subsection{Evaluation on graph-level tasks (RQ2)}
Besides node-level tasks, we further evaluate the performances of HCL and other baselines on graph classification under the linear evaluation protocol and answer the research question RQ2.

\noindent\textbf{Graph Classifications. }
\textbf{(1)} We compare our results with five \textbf{graph kernel methods} including shortest path kernel 
(SP) \cite{borgwardt_2005_icdm}, Graphlet kernel (GK) \cite{shervashidze_2009_ais}, Weisfeiler-Lehman sub-tree kernel (WL) \cite{shervashidze_2011_jmlr}, deep 
graph kernel (DGK) \cite{yanardag_2015_kdd}, and multi-scale Laplacian kernel (MLG) \cite{kondor_2016_nips} reported in \cite {sun2020infograph}. \textbf{(2)} We also compare 
with five \textbf{supervised GNNs} reported in \cite{xu_2019_iclr} including GraphSAGE \cite{hamilton_2017_nips}, GCN, GAT, and two variants of GIN: GIN-0 and 
GIN-$\epsilon$. \textbf{(3)} Moreover, We compare the results with other \textbf{unsupervised methods} including random walk \cite{gartner_2003_ltkm}, node2vec \cite{grover_2016_kdd}, sub2vec
\cite{adhikari_2018_pakdd}, graph2vec \cite{narayanan2017graph2vec}, InfoGraph\cite{sun2020infograph} , GCC\cite{qiu2020gcc}, GraphCL\cite{you2020graphcl} and MVGRL \cite{hassani2020contrastive}. The results shown in Table \ref{tab: graph_class} suggest that HCL achieves
superior results with respect to unsupervised models. For example, on REDDIT-B, HCL achieves 91.9\% accuracy, i.e., a 2.7\% 
relative improvement over previous state-of-the-art. 
When compared to supervised baselines individually, our model outperforms GCN and GAT models in 3 out of 5 datasets, e.g., a 
10.0\% relative improvement over GAT on IMDB-M dataset. 

Noting that HCL achieve superior and competitive performance on both node-level and graph-level tasks using a unified framework, unlike previous unsupervised models \cite{velickovic_2019_dgi,sun2020infograph}, we do not devise a specialized encoder for each task.

\subsection{Components Analysis and Ablation of HCL (RQ3 \& RQ4)}

Due to computation complexity, we conduct the ablation studies of proposed HCL on node classification of Cora and Citeseer datasets. All the experiment details are the same as mentioned in section 4.1. for fair comparison.

\begin{table}[!h]
\centering
\resizebox{0.8\columnwidth}{!}{%
\begin{tabular}{lcccc}
\hline
\textbf{ }    & Multi-Scale &   Multi-Channel         & {Cora}       & {Citeseer}   \\ \hline
HCL  & \checkmark & \checkmark  & \textbf{83.7 ± 0.6} & \textbf{73.3 ± 0.4} \\
HCL   & \checkmark & - & 83.4 ± 0.7 & 72.9 ± 0.5 \\
HCL   & - & \checkmark  & 83.0 ± 0.9 & 72.4 ± 0.7 \\ \hline
\end{tabular}
}

\vspace{0.5cm}

\caption{Ablation study of main components in HCL on Cora and Citeseer.}
\label{tab:ablation}
\end{table}

\noindent\textbf{Effect of Multi-scale and Multi-channel Contrastiveness (RQ3). }
To validate the effectiveness of the two contrastive components (Multi-scale, Multi-channel), We use HCL with/without multi-channel and multi-scale to denote the ablated model with one of the key components removed. The experiments on Cora and Citeseer presented in Table \ref{tab:ablation} show that HCL with both components yielded best performance, which demonstrates the effectiveness of our two contrastive schemes. 
 Specifically, the relative improvements are fair to be prominent as: multi-scale \& multi-channel, multi-scale, multi-channel are 2.2\%, 1.8\% and 1.3\% on Cora, 2.1\%, 1.5\% and 0.8\% on Citeseer, respectively ( HCL without multi-scale and multi-channel can be considered as DGI with Diffusion matrices as input, it yielded only 81.9 on Cora and 71.8 on Citeseer). 
These improvements can be attributed to the comprehensive multi-scale and multi-channel contrastive learning scheme, which takes the advantage of more flexible contrastiveness and more sufficient feature exploration.

\vspace{0.5cm}

\noindent\textbf{Effect of Pooling Settings (RQ4). }
To validate whether the multi-scale representation is useful at each of its scales in HCL, we conduct experiments on different scale settings. In the above part of Table \ref{tab:layers}, the experimental results suggested that removing scales decreased the graph learning performances. Each scale benefits from more multiplex self-supervision signals and empowered them to regularize each other. Moreover, we validate the advantages of the proposed L2Pool method on node classification task. We investigate three implementations for graph pooling methods: the proposed L2Pool, previous methods gPool~\cite{gao2019graph} and SAGPool~\cite{lee2019self}. As shown in the below part of Table \ref{tab:layers}, the experiments indicate that L2Pool yields superior performance, demonstrating more effective and proper scoring functions of adaptive L2Pool enables constructing more reasonable multi-scale graphs, via reducing the size of a graph while maintaining essential properties.

\begin{table}[!h]
\centering
\resizebox{0.75\columnwidth}{!}{%
\begin{tabular}{lccc}
\hline
Pooling-settings         & {Cora}       & {Citeseer}   \\ \hline
$HCL$ (4 scales: 1.0-0.9-0.8-0.7) & {83.7 ± 0.6}  & {73.3 ± 0.4} \\
$HCL$ (3 scales: 1.0-0.9-0.8) & 83.5 ± 0.8 & 73.0 ± 0.6 \\
$HCL$ (2 scales: 1.0-0.9) & 83.2 ± 0.7 & 72.8 ± 0.5 \\
$HCL$ (1 scales: 1.0) & 83.0 ± 0.9 & 72.4 ± 0.7  \\ \hline
$HCL_\emph{L2Pool}$  & \textbf{83.7 ± 0.6} & \textbf{73.3 ± 0.4} \\
$HCL_\emph{gPool}$ & 83.1 ± 0.7 & 72.5 ± 0.3 \\
$HCL_\emph{SAGPool}$ & 82.6 ± 0.8 & 72.2 ± 0.5 \\ \hline
\end{tabular}
}

\vspace{0.5cm}

\caption{Ablation study of pooling scales and methods in HCL.}
\label{tab:layers}
\end{table}

\begin{figure*}[!ht]
\centering
\includegraphics[width=0.9\columnwidth]{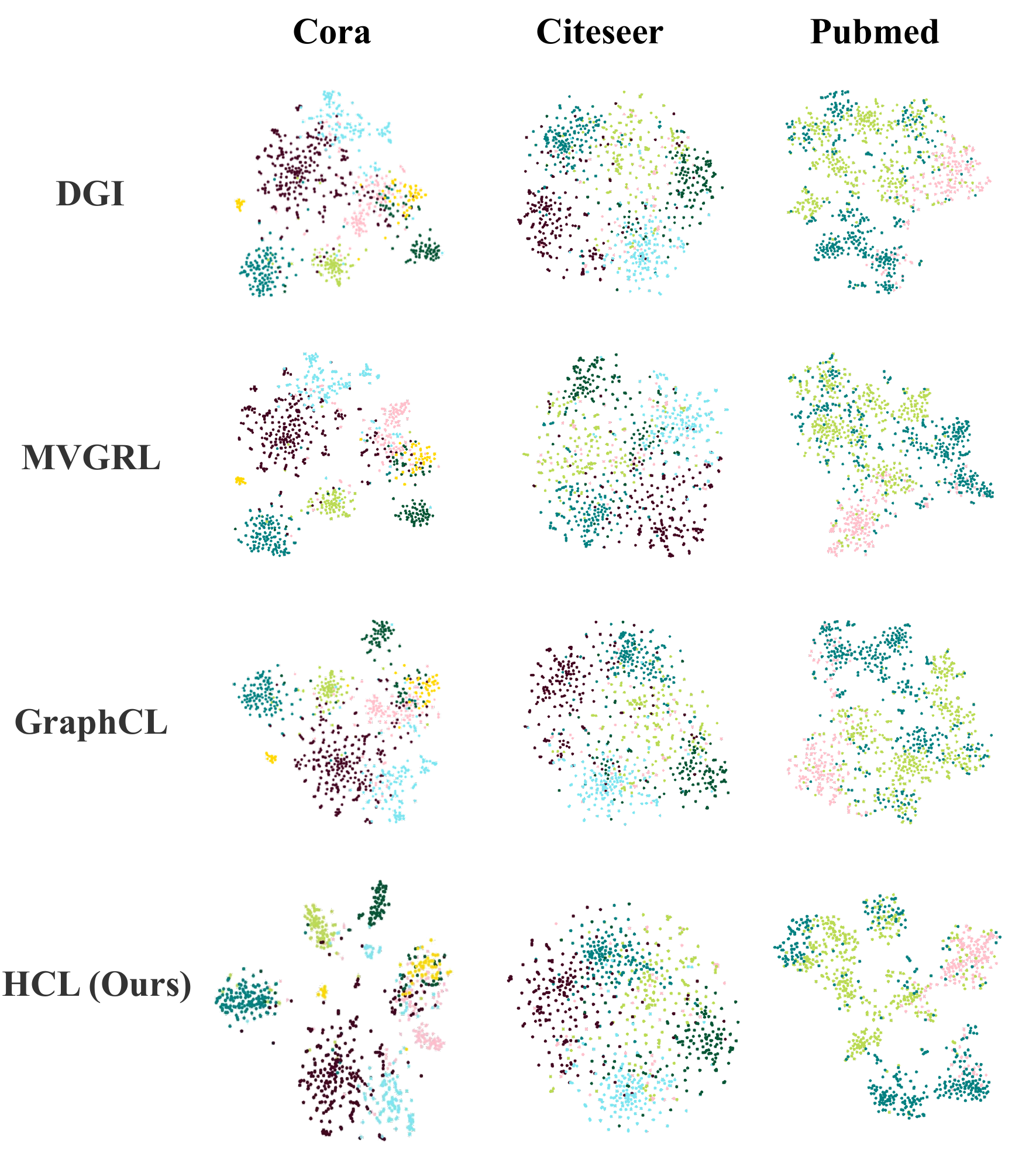}
\caption{t-SNE visualization of representation learned from different methods on Cora, Citeseer and Pubmed datasets.}
\label{vis}
\end{figure*}

\subsection{Further Analysis of Explainable Representation Visualization (RQ5)}
In this subsection, we further investigate the power of HCL to provide insightful interpretations and produce representation with prominent patterns in different graphs and answer research question RQ5. As shown in Figure \ref{vis}, we visualize the node embeddings of Cora, Citeseer and Pubmed calculated by different baselines via the t-SNE algorithm. Our HCL exhibits a relatively more compact and discernible clustering than other baselines, like DGI\cite{velickovic_2019_dgi}, MVGRL\cite{you2020graphcl} and GraphCL\cite{you2020graphcl}. It suggests that the hierarchical contrastive learning scheme of HCL captures more meaningful and interpretable clusters, which provides high-quality representations for the downstream tasks. To our knowledge, most previous methods neglected to capture the hierarchical structure, hindered by operating on a fixed-size scale. HCL is the first to explicitly integrate the hierarchical node-graph contrastive objectives in multiple-granularity, demonstrating superiority over previous methods.


\section{Conclusions}
In this work, we proposed a novel Hierarchical Contrastive Learning (HCL) framework for graph to explore more multiplex self-supervision signals and empowered them to regularize each other.
Extensive experiments suggest that (i) HCL outperforms most state-of-the-art unsupervised learning methods on node classification, node clustering and graph classification tasks; (ii) the proposed L2Pool methods yield more reasonable graph hierarchy with learnable topology-enhanced multi-head attention scores; (iii) the nested contrastive objective across multi-scale and multi-channel leads to better performances. 
Therefore, HCL paves the way to a potential direction for unsupervised graph learning objective and superior architecture design. In particular, the composite multi-scale and multi-channel contrastive objective bridges the gap between prior contrasting and hierarchical representation learning objectives, hence introduces a more sufficient and effective graph mining. 
In the future, the proposed HCL framework could be effectively integrated with more GNN models and applied on more graph learning tasks, to explore richer feature interaction for intrinsic informative pattern capturing.


%
%
%
\bibliographystyle{splncs03}
\bibliography{reference}

\end{document}